\documentclass[conference]{IEEEtran}
\IEEEoverridecommandlockouts
\usepackage{cite}
\usepackage{amsmath,amssymb,amsfonts}
\usepackage{algorithmic}
\usepackage{graphicx}
\usepackage{textcomp}
\usepackage{xcolor}
\usepackage{pdfpages}
\usepackage{arydshln}
\usepackage{booktabs}
\usepackage{authblk}
\usepackage{hyperref}
\hypersetup{
    colorlinks=true,
    linkcolor=blue,
    filecolor=magenta,      
    urlcolor=blue,
    pdftitle={Overleaf Example},
    pdfpagemode=FullScreen,
    }

\def\BibTeX{{\rm B\kern-.05em{\sc i\kern-.025em b}\kern-.08em
    T\kern-.1667em\lower.7ex\hbox{E}\kern-.125emX}}
    
\newcommand{\block}{\textbf}

\begin{document}

\title{Asymmetrical Bi-RNN for pedestrian trajectory encoding
}

\author[1,2*]{Raphaël Rozenberg \thanks{* This work was done during an internship at Centre de Robotique.}}
\author[2,3]{Joseph Gesnouin}
\author[2]{Fabien Moutarde}
\affil[1]{École Normale Supérieure - Université PSL, 75005 Paris, France}
\affil[2]{Centre de Robotique, MINES ParisTech - Université PSL,  75006 Paris, France}
\affil[3]{Institut Vedecom, 78000 Versailles, France}


\maketitle

\begin{abstract}

Pedestrian motion behavior involves a combination of individual goals and social interactions with other agents. In this article, we present an asymmetrical bidirectional recurrent neural network architecture called U-RNN to encode pedestrian trajectories and evaluate its relevance to replace LSTMs for various forecasting models. Experimental results on the Trajnet++ benchmark show that the U-LSTM variant yields better results regarding every available metrics (ADE, FDE, Collision rate) than common trajectory encoders for a variety of approaches and interaction modules, suggesting that the proposed approach is a viable alternative to the \textit{de facto} sequence encoding RNNs.
Our implementation of the asymmetrical Bi-RNNs for the Trajnet++ benchmark is available at: \href{http://github.com/JosephGesnouin/Asymmetrical-Bi-RNNs-to-encode-pedestrian-trajectories}{github.com/JosephGesnouin/Asymmetrical-Bi-RNNs-to-encode-pedestrian-trajectories}
\end{abstract}


\begin{figure*}[t]
\centerline{\includegraphics[width=\linewidth]{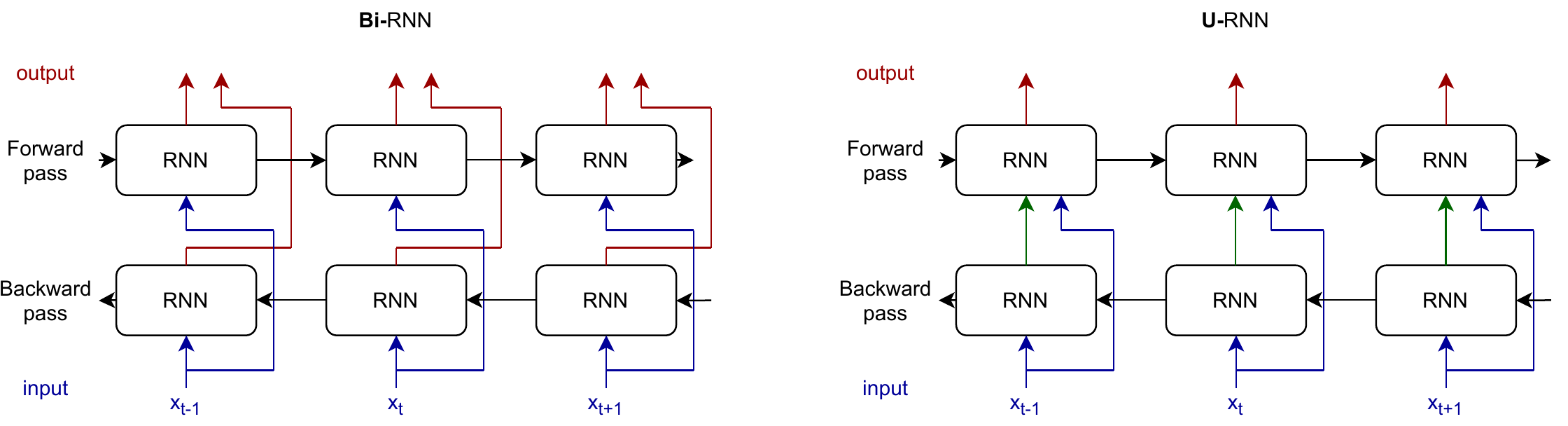}}
\caption{Comparison between Bi-RNN and U-RNN architectures (blue: inputs - red: outputs - black: hidden states - green: intermediate output).
U-RNN can use the information from the future during the forward pass, whereas the Bi-RNN only concatenates two naive readings in both directions.}
\label{ugru}
\end{figure*}

\section{Introduction}

\subsection{Pedestrian trajectory forecasting}

Pedestrian trajectory prediction from past positions using social interactions has been steadily receiving attention by the research community, as it plays a crucial role in various applications leading to the deployment of intelligent transport systems.

Following the success of Social LSTM \cite{alahi_social_2016} in trajectory forecasting in crowded scenes, a variety of approaches has been proposed that focused on efficiently leveraging social interactions from a scene \cite{gupta_social_2018,kothari_human_2021,bartoli2018context,pfeiffer2018data,vemula2018social,ma2016artificial}. In this article, we elude the question of improving social interactions models, and focus on the encoding of the trajectories of individual pedestrians by using U-RNNs (our asymmetrical Bi-RNNs) instead of regular LSTMs.

Using the recent Trajnet++ benchmark \cite{kothari_human_2021} and with respect to various available learning architectures that forecast pedestrians trajectories, we evaluate the effectiveness of U-RNNs for efficient pedestrian trajectories encoding. We then provide insight into designing improved motion encoders prior to the application of interaction modules for the task of pedestrian trajectory prediction.

\subsection{From Bi-RNNs to U-RNNs}

U-RNN is a bidirectional recurrent neural network architecture that was informally introduced in \cite{kaggle} under the form of U-GRUs for Knowledge Tracing. The objective of this work is to investigate whether U-RNNs could replace regular RNNs or Bi-RNNs for trajectory encoding.

Bi-RNNs \cite{650093} address a drawback of Recurrent Neural Networks (RNNs), which is that they cannot take the future into account when they encode an input, which may be desirable \cite{Goodfellow-et-al-2016} for some cases. 
For example, in the case of pedestrian trajectory prediction \cite{xue2017bi,yao2021bitrap}, one could expect that some movements are influenced by anticipation of a potential obstacle. Bi-RNNs produce two outputs, one that is obtained by reading the input forward and one by reading the input backwards. Concatenation or some other operation is then applied.

However, an aspect of Bi-RNNs that could be undesirable is the architecture's symmetry in both time directions. Bi-RNNs are often used in natural language processing, where the order of the words is almost exclusively determined by grammatical rules and not by temporal sequentiality. However, in trajectory prediction, the data has a preferred direction in time: the forward direction. Another potential drawback of Bi-RNNs is that their output is simply the concatenation of two naive readings of the input in both directions. In consequence, Bi-RNNs never actually read an input by knowing what happens in the future. Conversely, the idea behind U-RNN, illustrated in Fig.~\ref{ugru}, is to first do a backward pass, and then use during the forward pass information about the future. By using an asymmetrical Bi-RNN to encode pedestrian trajectories, we accumulate information while knowing which part of the information will be useful in the future as it should be relevant to do so if the forward direction is the preferred direction of the data.



\section{Related work}

\subsection{Encoder-Interaction-Decoder pipeline}

The most common pipeline for pedestrian trajectory prediction consists of:

\begin{enumerate}
    \item A \textbf{sequence encoder} for the past coordinates of each pedestrian independently.  The encoder is usually a RNN, such as a LSTM.
    \item An \textbf{interaction module} for taking into account the neighbors trajectories. The most common way to take into account the effect of interactions between agents in their trajectories is to decode the past positions while pooling on a spatial grid with either the neighbors' positions, their relative velocities \cite{kothari_human_2021}, or their RNN hidden states \cite{alahi_social_2016}. This last approach is very popular and is known under the name of \textit{social pooling}. 
    \item A \textbf{decoder} that predicts future coordinates. A common approach is to use a RNN for decoding. Some authors found that this can lead to error accumulation, and that a simple multi-layer perceptron (MLP) that predicts simultaneously all future positions performs better \cite{leal-taixe_red_2019}. However, taking into account interactions between pedestrians requires to predict the coordinates one step at a time, so RNNs are generally preferred.
\end{enumerate}

Most of past years research focused on improving the interaction module, with only limited new methods since \textit{Social-LSTM} \cite{alahi_social_2016}, or in developing approaches that take inspiration in popular frameworks such as Transformers \cite{giuliari_transformer_2020} or contrastive learning \cite{liu_social_2020} in order to deter the model from predicting colliding or too uncomfortable trajectories. However, little work has been published on the influence of the encoder and thus on the importance of past coordinates, even if it would be easily applicable on all models that use this pipeline.

\subsection{Alternative approaches.}

\paragraph{Learning-free algorithms.} The straight line at constant speed using the last known velocity is a reasonable approximation for the problem at hand \cite{scholler_what_2020}, given that we only try to predict the next few seconds. More complex learning-free methods can also be successfully applied, some generic, such as the Kalman Filter, and some specific, such as Optimal Reciprocal Collision Avoidance (ORCA) \cite{orca}, which ensures that trajectories do not collide, which is not necessarily the case with other methods, especially the straight line.
 
\paragraph{Other methods.} Even though non-RNN methods cannot take advantage of the research on interaction modules, alternative machine learning approaches have been developed. Convolutional Neural Networks are faster than RNN-based methods due to parallelization, but the performances are significantly lower \cite{nikhil_convolutional_2018}. Some authors have explored the popular Transformers architecture, but the results are inferior to those of RNNs with state-of-the-art social interaction modules \cite{giuliari_transformer_2020}. Research has also been conducted on applying Inverse Reinforcement Learning (IRL) to the pedestrian trajectory prediction problem \cite{fernando_neighbourhood_2019}, even though retrieving the pedestrian cost function requires much more computation than learning a predictor.

\begin{figure}[t]
\centerline{\includegraphics[width=0.7\linewidth]{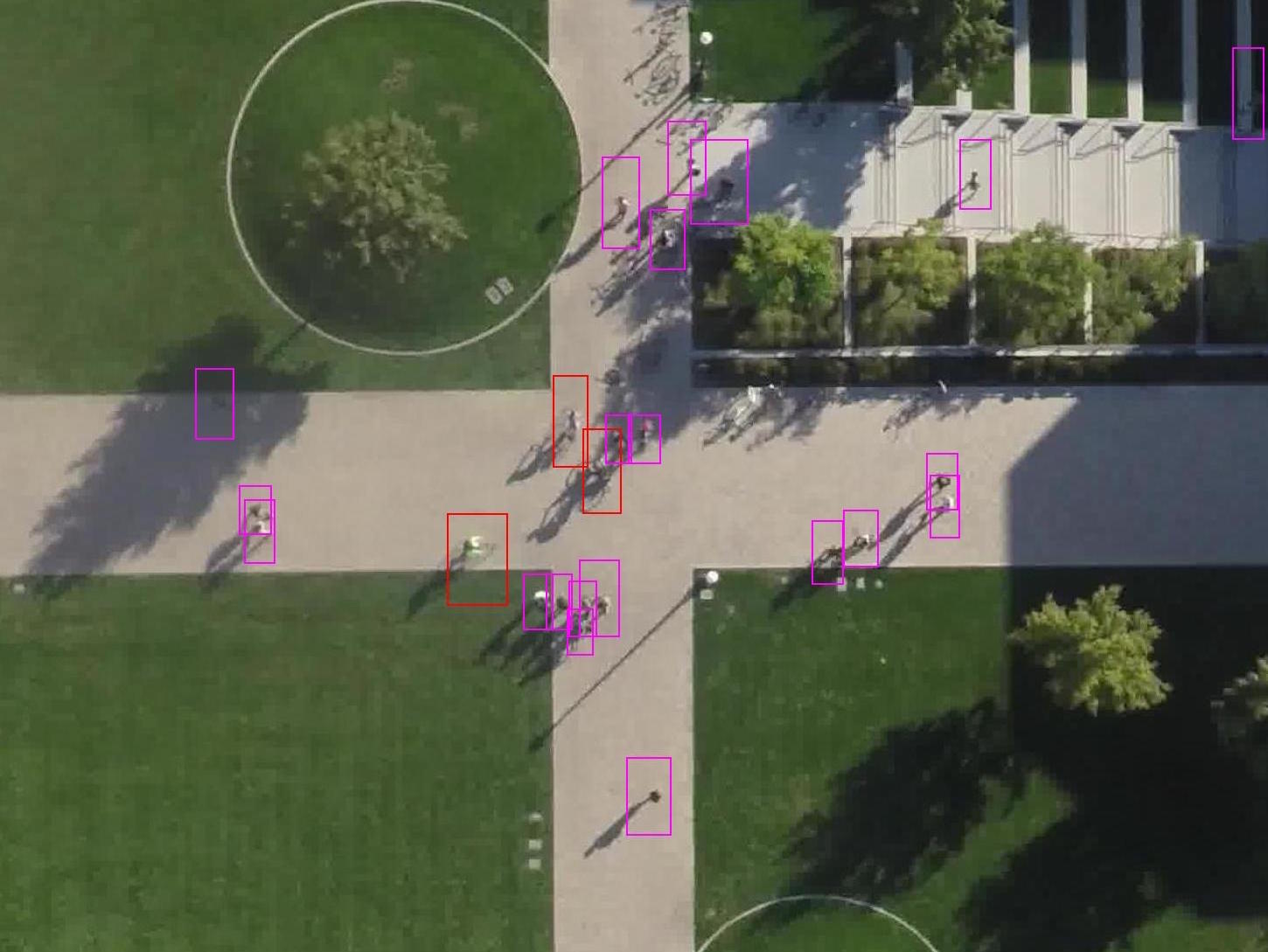}}
\caption{Sample from the Stanford Drone Dataset (which is not included in the Trajnet++ benchmark). The environment would play an important role in order to predict trajectories that do not go on the lawn.}
\label{stanford}
\end{figure}

\begin{figure*}[ht]
\centerline{\includegraphics[width=.32\textwidth]{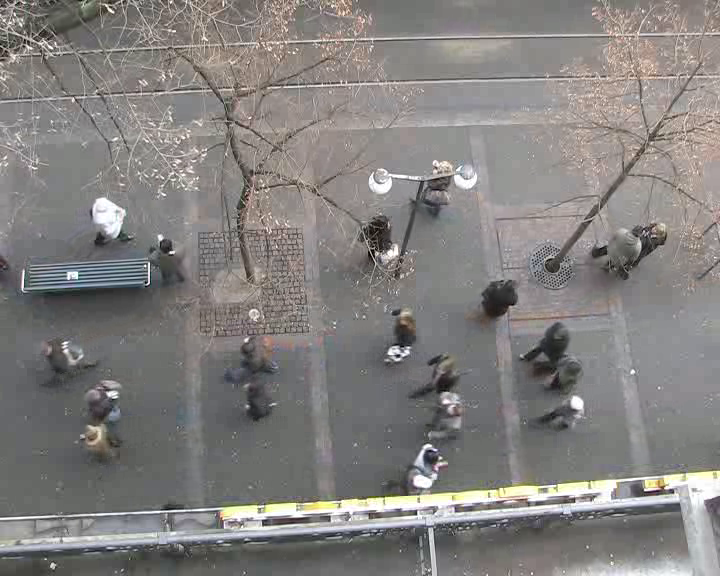}\hfill
\includegraphics[width=.32\textwidth]{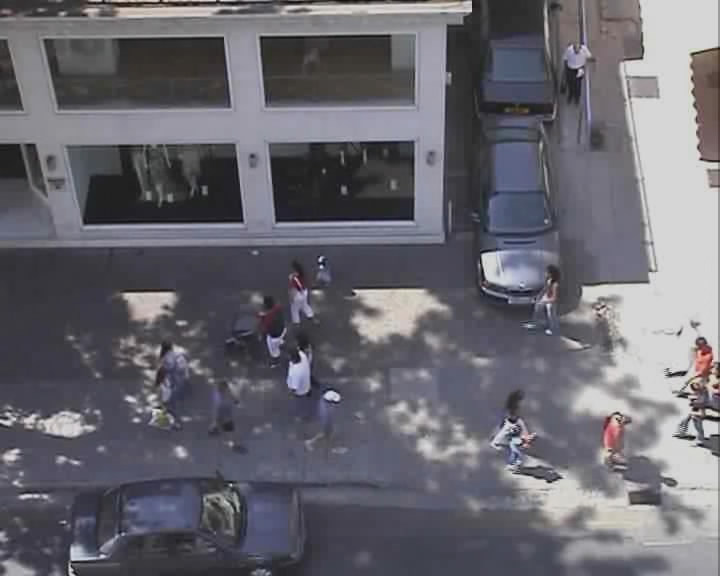}\hfill
\includegraphics[width=.32\textwidth]{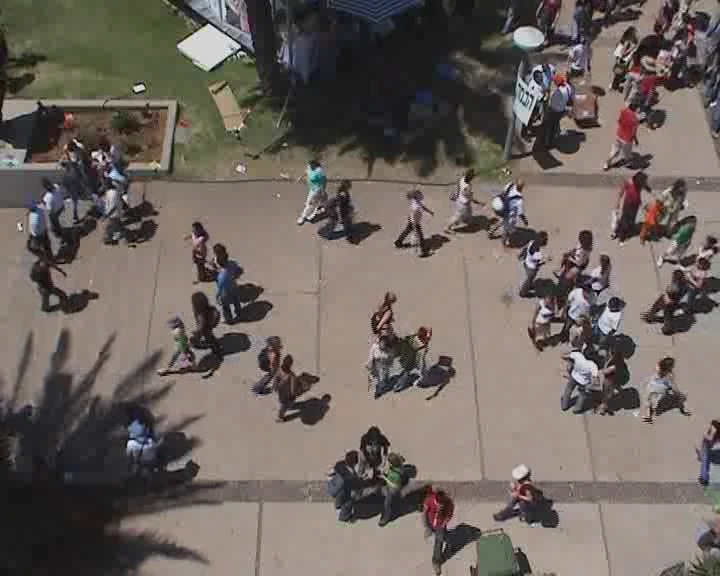}}
\caption{Images from different datasets from which the Trajnet++ benchmark trajectories are extracted. Left: ETH-hotel dataset - Center: UCY-zara dataset - Right: UCY-students dataset.}
\label{datasets}
\end{figure*}

\subsection{What information is relevant?} 

\paragraph{Scene context as an additional modality.} The Trajnet++ dataset does not include the pedestrians' environment, but some argue that it is sometimes necessary in order to predict trajectories correctly \cite{leal-taixe_red_2019}. Indeed, in situations such as the one in Fig.~\ref{stanford}, it would be very difficult to predict plausible trajectories since the environment would play an important role in order to predict trajectories that do not go on the lawn. However, the environment's additional information seems to make generalization more difficult \cite{scholler_what_2020}.

\paragraph{Neighbors past coordinates.}  Most methods make use of neighbors past and present positions. However, it seems that knowing even future neighbors positions is useless in terms of prediction error \cite{scholler_what_2020}. Indeed, global trajectories are not that much affected by interactions. Still, neglecting the influence of neighbors inevitably leads to collisions: relevant metrics for pedestrian trajectory prediction take this into account in addition to purely spatial errors, in order to produce physically feasible trajectories.

\section{Methodology}
\label{section:methodology}

We based our experiments on the Trajnet++ LSTM baseline \cite{kothari_human_2021} with respect to a variety of interaction modules: \textit{directional}, \textit{occupancy} and \textit{social} pooling. All hyper-parameters except for the encoder remained unchanged. For clarification purposes, we further explain our methodology for the \textit{directional} pooling case.

\subsection{Input embedding}

The input data consists of coordinates $(x_t)_{t \in [\![1, T_{obs}]\!]}$ for each pedestrian. In order to allow easier generalization, we use velocities $(v_t)_{t \in [\![1, T_{obs}-1]\!]}$ instead with $v_t=x_{t+1}-x_t$.

From the trajectory velocities $(v_t)_t$ of a single pedestrian, we obtain the trajectory embeddings $(e_t)_{t \in [\![1, T_{obs}-1]\!]}$ with 
$$e_t = f(v_t, W_e)$$
where $f$ is a single-layer perceptron, and $W_e$ are learnable weights that are shared among pedestrians.

\subsection{U-RNN architecture}

The backward and forward hidden states $(h^b_t)_{t \in [\![1, T_{obs}-1]\!]}$ and $(h^f_t)_{t \in [\![1, T_{obs}-1]\!]}$ are obtained according to these equations:

\begin{IEEEeqnarray*}{rrCl}
& h^b_{t-1} & = & RNN(h^b_t, e_t, W_b)\\*
& h^f_{t+1} & = & RNN(h^f_t, [e_t, h^b_t], W_f)\\*
\end{IEEEeqnarray*}

where $W_b$ and $W_f$ are learnable weights that are shared among pedestrians, and $[\cdot, \cdot]$ denotes concatenation.

The last hidden state $h^f_{T_{obs}}$ is then used as the encoding of the sequence.

\subsection{Decoder}

For decoding, we used a RNN and directional pooling, with a learnable $GridPooling$ function that involves average pooling and a linear embedding, all of which we do not detail here and was implemented in \cite{kothari_human_2021}. The predicted positions $(o^i_t)_{t \in [\![1, T_{pred}]\!]}$ of pedestrian $i$ are obtained according to these equations:

\begin{IEEEeqnarray*}{rrCl}
& h^{i}_1 & = & h^{f, i}_{T_{obs}}\\*
& e^i_t & = & f(v^i_t, W_e) \\*
& I^i_t& = & GridPooling(v^{-i}_t) \\*
& h^{i}_{t+1} & = & RNN(h^{i}_t, [e^i_t, I^i_t], W_d)\\*
& o^i_t & = & g(h^i_t, W_{out}) \\*
\end{IEEEeqnarray*}

where $(v^i_t)_{t}$, $(e^i_t)_{t}$, $(I^i_t)_{t}$, $(h^i_t)_{t}$ are respectively the velocities, velocity embeddings, interaction embeddings and decoder hidden states for pedestrian $i$, $W_e$, $W_d$ and $W_{out}$ are learnable weights that are shared among pedestrians ($W_e$ being the same as for the encoder), $v^{-i}$ denotes velocities of pedestrians other than $i$ and $[\cdot, \cdot]$ denotes concatenation.

\section{Experiments}

\begin{figure*}[ht]
\centerline{\includegraphics[width=0.7\linewidth]{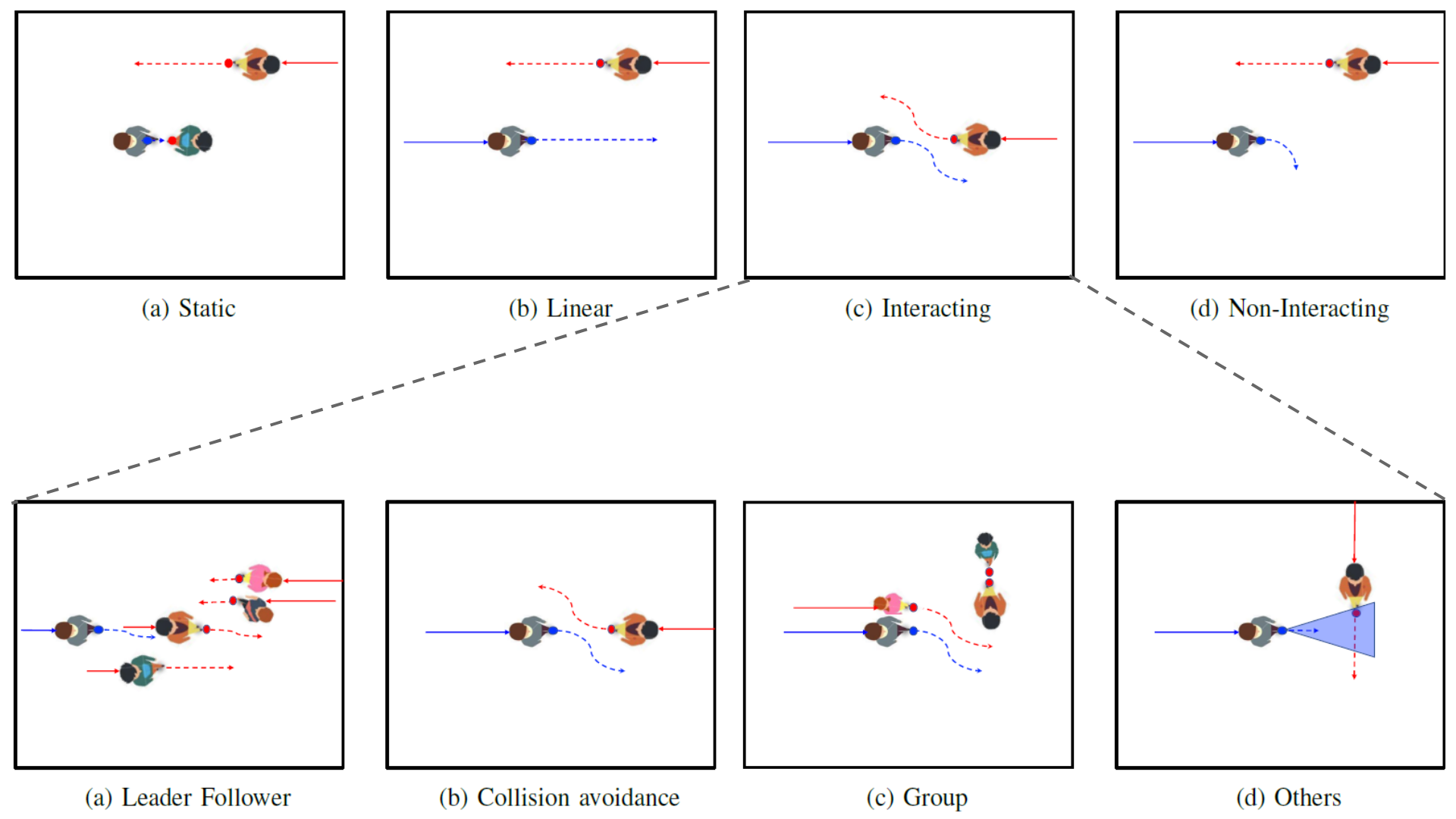}}
\caption{Classification of trajectories in Trajnet++ according to the interactions between agents. Visualization from \cite{kothari_human_2021}.}
\label{trajectories}
\end{figure*}

\subsection{The Trajnet++ benchmark}

There are several datasets that are aimed at evaluating pedestrian motion prediction, with very diverse characteristics~\cite{rudenko_human_2020}.
We chose the Trajnet++ benchmark \cite{kothari_human_2021}, that aggregates several common pedestrian trajectories datasets, emphasis the importance of quantifying the physical feasibility of a model prediction and only evaluates on trajectories where there are interactions between pedestrians. 

\block{Data.} Trajnet++ data consists of trajectories that have been extracted from real-life videos and that are under the form of spatial coordinates. The framerate is 2.5 frames per second.
The datasets that are used are:
\begin{itemize}
    \item ETH \cite{pellegrini2010improving}, itself subdivided into ETH-hotel and ETH-uni. $\sim$650 tracks extracted from 25 min of video.
    \item UCY \cite{leal2014learning}, itself subdivided into UCY-zara and UCY-students. $\sim$700 tracks extracted from 16 min of video.
    \item WildTrack \cite{chavdarova2018wildtrack}, $\sim$650 tracks extracted from an hour of video.
    \item L-CAS \cite{sun20183dof}, $\sim$1100 tracks extracted from 49 min of video.
    \item CFF \cite{alahi2014socially}, Large-scale dataset of $\sim$42 million trajectories extracted from real-world train stations.
\end{itemize}
Fig.~\ref{datasets} illustrates ETH and UCY datasets with sample images from videos from which the spatial coordinates were extracted.

In addition, synthetic data generated using ORCA \cite{orca} is also used.

\block{Task.} The goal is to predict the spatial coordinates of pedestrians in the near future (12 frames, i.e. 4.8 seconds), using only the near past (9 frames, i.e. 3.6 seconds). In each scene (set of different agents' trajectories over a given duration), a primary pedestrian is designated for evaluation purposes.
 
\block{Categories.} The scenes in the data are subdivided into categories with respect to the primary pedestrian of the scene, as Fig.~\ref{trajectories} illustrates. Type~I and Type~II denote respectively static primary pedestrian trajectories and trajectories that are correctly predicted with an extended Kalman filter. Type~III is the benchmark's type of interest, as it regroups all scenes where the primary pedestrian has interactions with other agents. Type~IV is used for the remaining scenes, where the primary pedestrian trajectory seems unpredictable even when given the social environment. 

In addition to the four main types, Type~III is further subdivided into four categories that describe the main type of interaction that is occurring: Leader-follower (the primary pedestrian follows someone else), Collision avoidance (the primary pedestrian had to avoid someone else), Group (the primary pedestrian is part of a group) and Others.

\block{Metrics.} There are four main metrics. Two are spatial errors: Average Displacement Error (ADE) and Final Displacement Error (FDE), that are expressed in meters. The other two are collision errors: Prediction Collision (Col-I) and Ground Truth Collision (Col-II), that are expressed in percentage. Col-I is the fraction of collisions between the primary pedestrian predicted trajectory and the other pedestrians predicted trajectories, and thus represents how physically realistic the predicted scene is, regardless of reality. Col-II, on the other hand, is the fraction of collisions between the primary pedestrian predicted trajectory and the other pedestrians \textit{real} trajectories. Therefore, it represents how physically realistic the predictions are individually.

\block{Evaluation.} According to the Trajnet++ benchmark, the performance is evaluated on $\sim$3000 scenes from ETH and UCY datasets, as well as on $\sim$4000 synthetic scenes. The benchmark gives metrics for each type and sub-type of scene. The score that is chosen in order to compare models on the public leaderboard is FDE computed on Type III (Interacting) scenes from the real datasets ($\sim$1700 scenes), with Col-I as the secondary score (computed on the same data). Until the end of March 2021, the secondary score was FDE computed on Type III scenes from the synthetic dataset, but it was abandoned because predicting synthetic trajectories had become a solved problem. On the contrary, while performances seem to have reached a limit with respect to FDE (more than one meter on a 4.8 seconds horizon), the current challenge is to be able to predict physically feasible scenes while keeping a good FDE.

\setlength\tabcolsep{4pt}
\begin{table}[ht]
\caption{Results for several baselines and for the best submission on the Trajnet++ public leaderboard (with respect to FDE).}
\begin{tabular}{c|c|c|c|c}
\toprule
\multicolumn{1}{c}{\textbf{Model}}    & \multicolumn{1}{c}{\textbf{ADE (m)}} & \multicolumn{1}{c}{\textbf{FDE (m)}} & \multicolumn{1}{c}{\textbf{Col-I (\%)}} & \textbf{Col-II (\%)}  \\ \hline
Kalman filter                & 0.87          & 1.69            & \textbf{0 }                  & 19.5       \\ \hdashline
Constant velocity  \cite{scholler_what_2020}         & 0.68          & 1.42            & 14.3               & 15.2       \\ \hdashline
ORCA \cite{orca}                         & 0.72          & 1.42            & \textbf{0}                   & \textbf{11.3}       \\ \hdashline
Vanilla LSTM                 & 0.67          & 1.43            & 15.2               & 12.3       \\ \hdashline
AMENet \cite{cheng2021amenet}    & 0.62         & 1.30            & 14.1                &  16.9       \\
\hdashline
AIN \cite{zhu2020robust}     & 0.62         & 1.24            & 10.7                &  17.1       \\
\hdashline
PecNet \cite{mangalam2020not}     & 0.57          & 1.18            & 15.0                &  14.3       \\
\hdashline
Social NCE \cite{liu_social_2020} & \textbf{0.53}          & \textbf{1.14}            & 5.3                & \textbf{11.3 }      \\
\bottomrule
\end{tabular}
\label{tab1}
\end{table}

\setlength\tabcolsep{6pt}
\begin{table*}[t]
\caption{Comparison of motion-encoding designs with respect to various interactions modules architectures on interacting trajectories of TrajNet++ real world dataset.}
\begin{center}
\scalebox{0.95}{
\begin{tabular}{c|c|c|c|c|c}
\toprule
\multicolumn{1}{c}{\textbf{Model}}    & \multicolumn{1}{c}{\textbf{Interaction}} & \multicolumn{1}{c}{\textbf{ADE (m)}} & \multicolumn{1}{c}{\textbf{FDE (m)}} & \multicolumn{1}{c}{\textbf{Col-I (\%)}} & \textbf{Col-II (\%)}  \\
\multicolumn{1}{c}{\textbf{(Encoder - Decoder)}} & \multicolumn{1}{c}{}                     & \multicolumn{1}{c}{± 0.01 m} & \multicolumn{1}{c}{± 0.01 m} & \multicolumn{1}{c}{± 0.5\%}  & ± 1\%    \\ \hline
Constant velocity \cite{scholler_what_2020} & None             & 0.68          & 1.42            & 14.3               & 15.2       \\ \hdashline
None - GRU        & Dir. pooling \cite{kothari_human_2021}    & 0.63          & 1.33            & 6.9                & 12.1       \\ \hdashline
LSTM - LSTM     & Occ. pooling \cite{alahi_social_2016}    & 0.58          & 1.23            & 11.5                &  13.9       \\
\textbf{U-LSTM - LSTM}     & Occ. pooling \cite{alahi_social_2016}    & \textbf{0.57}          & \textbf{1.22}            & \textbf{10.2}                & 14.9 \\

\hdashline

GRU - GRU         & Dir. pooling \cite{kothari_human_2021}   & 0.58          & 1.24            & 6.5                 & 12.4      \\
Bi-GRU - GRU      & Dir. pooling \cite{kothari_human_2021}    & 0.59 & 1.26    & 6.7           & 11.7   \\ 
U-GRU - GRU       & Dir. pooling \cite{kothari_human_2021}    & 0.58 & 1.25     & 6.5           & 11.7   \\ 
reversed U-GRU - GRU       & Dir. pooling \cite{kothari_human_2021}    & 0.58 & 1.25  & 6.5           & \textbf{11.0}  \\ 
LSTM - LSTM     & Dir. pooling \cite{kothari_human_2021}    & 0.58          & 1.25            & 6.4                &  11.4       \\
Bi-LSTM - LSTM      & Dir. pooling \cite{kothari_human_2021}    & 0.59 & 1.28    & 6.2           & 11.9  \\
\textbf{U-LSTM - LSTM}     & Dir. pooling \cite{kothari_human_2021}    & \textbf{0.56}          & \textbf{1.22}            & \textbf{5.2}                & 11.9 \\
reversed U-LSTM - LSTM       & Dir. pooling \cite{kothari_human_2021}    & 0.58  & 1.26   & 6.6 & \textbf{11.1}   \\ \hdashline
LSTM - LSTM     & Soc. pooling \cite{alahi_social_2016}    & 0.55         & 1.18            & 6.9                &  12.7       \\
\textbf{U-LSTM - LSTM}     & Soc. pooling \cite{alahi_social_2016}    & \textbf{0.53}          & \textbf{1.15}            & \textbf{6.5}                & 11.5 \\

\hdashline
Social NCE \cite{liu_social_2020} & Soc. pool. \cite{alahi_social_2016} + contr. learning  & \textbf{0.53}          & \textbf{1.14}            & \textbf{5.3}                & 11.3       \\

\bottomrule
\end{tabular}
\label{tab2}}
\end{center}
\end{table*}

\subsection{Baselines}

We used the following baselines for comparison purposes:
\begin{itemize}
    \item \textbf{Learning-free methods.} We considered Kalman filter, constant velocity \cite{scholler_what_2020} and ORCA \cite{orca}.
    \item \textbf{Vanilla LSTM.} An architecture with a LSTM encoder, a LSTM decoder, and no interaction module (each pedestrian is considered independently).
    \item \textbf{AMENet \cite{cheng2021amenet}}, a conditional variational auto-encoder based on attentive dynamic maps for interaction modeling, \textbf{AIN \cite{zhu2020robust}}, an encoder-decoder pipeline focusing on global spatio-temporal interactions and \textbf{PecNet \cite{mangalam2020not}}, a conditioned-on-goal endpoint variational auto-encoder. We reference the scores that are on the public leaderboard for AMENet and the ones referenced in \cite{kothari_human_2021} for AIN and PecNet.
    \item \textbf{Social NCE \cite{liu_social_2020}.} Best submission on the public leaderboard, with respect to FDE. It uses social pooling and contrastive learning. We reference the scores that are on the public leaderboard.
\end{itemize}

Table~\ref{tab1} shows the results on the four metrics and helps understand the pros and cons of each method. In terms of FDE, the Kalman filter is by far the worst of all, almost 30~cm behind constant velocity (but Type III scenes, on which evaluation is performed, are by definition scenes where trajectories cannot be correctly predicted using a Kalman filter). The constant velocity method is both extremely simple and reasonably effective, but at the cost of high collision rates. ORCA allows to completely get rid of collisions without sacrificing FDE. Vanilla LSTM is completely irrelevant, since it is worse even than
the constant velocity method, highlighting how the potential of RNNs can only be revealed by using interaction encoders. Finally, the best submission on the leaderboard reaches a FDE that is 30~cm below the constant velocity method, with a Col-I of only 5\%; however, as we said, ADE and FDE are still relatively high in absolute terms.

\subsection{Experiments details}

For training, we used ETH, UCY, WildTrack, L-CAS, and only part of CFF datasets, totalling $\sim$29000 scenes in the training set and $\sim$5000 scenes in the validation set. In the training procedure, we decrease the learning rate when the validation loss reaches a plateau, and also apply early-stopping when the validation loss stops decreasing for several epochs. We also use rotation augmentation as a data augmentation technique to regularize all the models.

We did not code everything from scratch, but rather built on top of the numerous baselines that are available with Trajnet++. Since out goal was not to beat the state-of-the-art but rather to allow meaningful comparison between different motion encoders, comparisons of given approaches are relevant given the same interaction module and hyper-parameter settings.

We tested the following architectures, denoted by their Encoder-Decoder structure.
For each architecture, RNN can be replaced by either GRU \cite{cho2014properties,chung2014empirical} or LSTM \cite{hochreiter1997long} (we did not test combinations of both):
\begin{itemize}
    \item \textbf{RNN - RNN.} A common baseline.
    \item \textbf{Bi-RNN - RNN.} We used concatenation in order to fuse the outputs of the Bi-RNN, since it worked better than summation.
    \item \textbf{U-RNN - RNN.} The architecture described in Section~\ref{section:methodology}.
    \item \textbf{reversed U-RNN - RNN.} The backward pass and forward pass are inverted in the U-RNN, in order to investigate if there is indeed a preferred direction of U-RNNs according to the data.
\end{itemize}

We used default number of parameters that were similar to the baselines in \cite{kothari_human_2021} and did not change between different models. However, this led to LSTM models having higher total number of parameters than their GRU counterparts, but it did not affect our conclusions. The order of magnitude of the uncertainties on the metrics were ± 1 cm on ADE and FDE, ±~0.5\% on Col-I and ±~1\% Col-II.

\subsection{Results}

In Table~\ref{tab2}, we present the results that we obtained during our experiments. The first thing to notice is that using a simple RNN decoder with \textit{directional} pooling, even without an encoder, improved FDE by 10 cm and cuts Col-I by half compared to the Constant velocity model or to Vanilla LSTM. Secondly, adding a RNN encoder for past coordinates helped improving performance, which indicates that there is indeed relevant information in past positions. This suggests that pedestrians engage in complex trajectories that may span on relatively long durations.

Note that the proposed asymmetrical architecture is independent of the chosen recurrent unit. We observed in preliminary experiments that the encoder's architecture did not seem to have any impact, with identical performances of GRU - GRU, Bi-GRU - GRU, U-GRU - GRU and reversed U-GRU - GRU architectures. At first glance, one could conclude that the information contained in past coordinates may be too redundant to allow to detect any difference between encoder architectures, as there would be no further information to extract. Or that contrary to vehicles for example, pedestrian trajectories are too irregular to make good use of past information. However, experiments with LSTMs gave different results. LSTM~-~LSTM and Bi-LSTM - LSTM performed similarly as GRU architectures, but using a U-LSTM encoder helped get significantly better ADE, FDE and Col-I for \textit{directional} pooling, suggesting that there was indeed unused information in past trajectories. Regarding Col-II, the best architectures seem to differ compared to the other metrics, but this appears to be non-significant given the small score differences and the order of magnitude of the standard deviations.


The better performance of U-LSTM compared to U-GRU strongly indicates that the additional information extracted by the U-RNN architecture came from long-term dependencies. Moreover, the hypothesis we proposed, that the non-symmetrical architecture of U-RNN should better leverage information by using the preferred direction of the data is supported by the absence of performance improvement when using a reversed U-LSTM encoder.

Since it was clear that, for the \textit{directional} pooling case, the proposed Asymmetrical Bi-RNNs motion encoder performed better than regular LSTMs  which are the \textit{de facto} RNNs for trajectory encoding, we experimented U-LSTMs with \textit{occupancy} and \textit{social} pooling. In both experiments, our sequence encoder yielded significantly better results compared to regular LSTMs for every available metric (ADE,FDE, Col I). This suggests that the proposed architecture is a viable alternative to LSTMs for trajectory encoding.

\section{Conclusion}

In this article, we proposed a sequence encoder based on Asymmetrical Bi-RNNs to predict future pedestrians trajectories using naturalistic pedestrian scenes data from the widely studied Trajnet++ dataset. Contrary to many previous studies that proposed new interactions modules, our work solely relies on proposing a new sequence encoder that could easily be applied to all models that use the encoder-decoder pipeline for pedestrian trajectory forecasting, while taking advantage of the research on interactions and multi-modal trajectory prediction. The proposed sequence encoder was shown to achieve better prediction accuracy than previous sequence encoders such as LSTMs for a variety of existing approaches and interactions modules. This suggests that there is still room for improvement in coordinates-only approaches, and indicates that interactions are not the only aspect on which pedestrian trajectory prediction can progress.
Although this work is highly preliminary, our quantitative results open many perspectives for future research. The success of Asymmetrical Bi-LSTMs compared to Asymmetrical Bi-GRUs suggests that this boost may come from using information with long-term dependencies, confirming that some pedestrians movements are influenced by long-term anticipation. We believe that these results constitute a promising baseline to replace LSTMs for a variety of approaches and could be used to significantly improve current pedestrian trajectory prediction algorithms.

\bibliographystyle{IEEEtran} 
\bibliography{refs.bib}

\end{document}